\documentclass[12pt]{spieman}  

\usepackage{amsmath,amsfonts,amssymb}
\usepackage{graphicx}
\usepackage{bm}
\usepackage{mathrsfs}
\usepackage{booktabs}
\usepackage{multirow}
\usepackage{subfigure}
\usepackage[subfigure]{tocloft}
\usepackage{cite}
\usepackage{overpic,color}
\usepackage{setspace}
\usepackage{lineno}
\usepackage{enumerate}
\usepackage{textcomp}

\hypersetup{linkcolor=black, urlcolor=black,citecolor=black}

\newcommand{\figref}[1]{Fig.~\ref{#1}}
\newcommand{\tabref}[1]{Table~\ref{#1}}
\newcommand{\equref}[1]{Eq.~\ref{#1}}
\newcommand{\secref}[1]{Sec.~\ref{#1}}
\newcommand{\etal}{\textit{et~al.} }

\title{Saliency detection based on structural dissimilarity induced by image quality assessment model}
\author[a]{Yang Li}
\author[a]{Xuanqin Mou}
\affil[a]{Xi¡¯an Jiaotong University, Institute of Image Processing and Pattern Recognition, Xi¡¯an, China, 710049}

\cftpagenumbersoff{figure}
\cftpagenumbersoff{table}

\begin{document}
\maketitle

\begin{abstract}
\urlstyle{same}
The distinctiveness of image regions is widely used as the cue of saliency. Generally, the distinctiveness is computed according to the absolute difference of features. However, according to the image quality assessment (IQA) studies, the human visual system is highly sensitive to structural {changes} rather than absolute difference. Accordingly, we propose the computation of the structural dissimilarity between image patches as the distinctiveness measure for saliency detection. Similar to IQA models, the structural dissimilarity is computed based on the correlation of the structural features. The global structural dissimilarity of a patch to all the other patches represents saliency of the patch. We adopted two widely used structural features, namely the local contrast and gradient magnitude, into the structural dissimilarity computation in the proposed model. Without any post-processing, the proposed model based on the correlation of either of the two structural features outperforms 11 state-of-the-art saliency models on three saliency databases. { Source code of the proposed model can be downloaded at {\color{black}\url{https://github.com/yangli-xjtu/SDS}}}. \emph{This is the draft manuscript of our paper accepted by Journal of Electronic Imaging. doi: 10.1117/1.JEI.28.2.023025}

\end{abstract}


\begin{spacing}{2}

\section{Introduction}

The human visual system (HVS) can rapidly attend to salient regions from complex scenes.
Saliency model, which aims to mimic this mechanism and predict where a human looks in an image, has attracted extensive research interest in computer vision.
Regarding the human-fixation data 
{ under the free-viewing task when viewing images}
as the ground truth, the saliency models usually generate a saliency map with sparse blobs to predict the density of the fixation on the image.
Such a saliency map can be used to benefit many vision-related applications, such as image quality assessment \cite{zhang2017study}, video compression\cite{Itti2004}, and image retargeting\cite{Goferman2012a}.

A number of saliency models have been proposed in recent years. To detect the salient regions, inspired by the biological theories \cite{treisman1980feature,desimone1995neural}, many saliency models highlight image regions that stand out from their surroundings. The distinctiveness-based models are the most representative models in this regard. The center-surround absolute difference on the biologically motivated features was first proposed by Itti \etal \cite{Itti1998} and is widely used to detect the local distinctive regions \cite{fang2016learning,ishikura2018extreme}. Later, by proposing a graph model to globally integrate the image regions, Harel \etal showed that the global distinctiveness could better predict saliency. Further, the spatially weighted global absolute difference of image patches in the principal components space \cite{duan2011visual} or CIELab color space \cite{Goferman2012a,zhou2013new} also presents high accuracy in saliency detection.
{Besides, Rigas \etal \cite{rigas2015efficient} proposed to compute salience of the image patch by the distinctiveness based on the Hamming distance of sparse coding coefficients.}
By computing the probability of features, the local or global distinctive regions can also be effectively highlighted by the probability-based models \cite{Bruce2006saliency,tavakoli2011fast}. In addition, Hou \etal introduced the computation of saliency into the spectral domain. Although the spectral models usually have low complexity, they lack the biological plausibility \cite{Borji2013}. The learning-based models have recently become popular owing to the powerful data-driven deep learning technology \cite{vig2014edn,liu2015MrCNN,kruthiventi2017deepfix,liu2018deep,wang2018deep,he2018catches}. However, their generalization ability needs further exploration \cite{li2015finding}. Therefore, in this study, we focused on the distinctiveness-based models.

Moreover, another popular topic in computer vision, that is, perceptual image quality assessment (IQA) has also evoked great research interest in recent years. The goal of IQA is to predict the image quality consistent with the human subjective evaluation. When a pristine image is available, the IQA model type called the full reference IQA (FR-IQA) model computes the structural similarity between the given and pristine images, for example, the famous structural similarity metric (SSIM) \cite{wang2004image}. The underlying assumption is that the HVS is sensitive to the structural change when evaluating the image quality. Specifically, the structural similarity measure is usually a point-by-point correlation of the common features, such as the local contrast (LC) \cite{wang2004image} and gradient magnitude (GM) \cite{Xue2014}. As both saliency detection and FR-IQA reflect how the HVS perceives the difference between image regions \cite{Zhang2014}, the relationship between saliency and IQA has attracted much attention. In many IQA models, the saliency map is used to predict the importance of distortion to human perception. Specifically, the distortion in salient regions was proved to be perceived more annoying by Engelke \etal \cite{engelke2010linking}. Then, some studies adopted the saliency map as weight in the computation of average similarity~\cite{moorthy2009visual,tong2010full,farias2012performance,gu2016saliency,zhang2017study}. An interesting exception is the work proposed by Zhang \etal \cite{Zhang2014}, who adopted the saliency maps as the feature maps for comparing the similarity between images. Overall, the migration from saliency detection to IQA widely exists in literature. However, to the best of our knowledge, no investigations have yet been conducted on the exploitation of IQA models for saliency detection.

This paper contributes toward saliency detection by exploiting IQA models. Many previous saliency models computed the absolute difference of features to measure the distinctiveness of image regions in the local \cite{Itti1998,ishikura2018extreme} or global extent \cite{Harel2006,duan2011visual,Goferman2012a}. However, according to the studies on IQA, the structural similarity is validated as a better similarity measure than the absolute difference in the context of HVS perception. Therefore, we propose using the structural dissimilarity as the distinctiveness measure between image patches for saliency detection. Specifically, the structural dissimilarity is computed as one minus the correlation of structural features, i.e., the relative difference of structural features. Then, the global structural dissimilarities of a patch to all the other patches are {spatially weightedly} summed to represent saliency of the patch.
Besides, two widely used structural features, i.e, the LC and GM, extracted in the YIQ color space are adopted into the computation of the structural dissimilarity.

{
The proposed model has a similar framework (i.e., spatially weighted dissimilarity) to many other models \cite{duan2011visual,Goferman2012a,zhou2013new}.
In contrast to these models, the proposed model adopts a novel structural dissimilarity measure and intentionally employs a simple framework that excludes the post-processing or multi-scale operation.
Based on the validation on three databases (i.e., Toronto \cite{Bruce2006saliency}, ImgSal \cite{li2013visual}, and MIT \cite{Judd2009Learning}),
the proposed model outperforms 11 state-of-the-art saliency models.
The effectiveness of the proposed model and the adopted similar but simpler framework indicate the high relevance of the proposed structural dissimilarity and the allocation of fixation under the free-viewing task when viewing images, thereby highlighting the novelty of the proposed model.

Furthermore, 
the adopted LC in the YIQ color space is similar to
the center-surround contrast of the colors and intensity in Itti's model.
However, Itti \etal summed the maximum-mean-local-maxima (MMLM) normalized feature maps, whereas we summed the patch-based global relative differences of features as the salience.
The better performance of the proposed model implies that the proposed method would be a better normalization approach for highlighting the salient regions.
A further experimental comparison verifies this claim.

Moreover, a comprehensive analysis was conducted to study the effectiveness of the proposed model in terms of three aspects: the spatially weighted dissimilarity-based framework, the structural features, and the correlation-deduced relative difference measure.

The contributions of this paper are summarized as follows:
1) Structural dissimilarity is introduced into saliency detection for the first time.
2) The proposed model is highly competitive with the state-of-the-art models.
3) The global structural dissimilarity is more effective in highlighting salient regions than the widely used normalization approaches.
4) A comprehensive analysis was conducted to better understand the reasons for the good performance of the proposed model.

}

The remainder of the paper is organized as follows.
Section~\ref{sec:relate} introduces the related work and background.
Particularly, a brief introduction of IQA is presented to bridge the knowledge gap between saliency detection and IQA.
Section~\ref{sec:methods} presents the proposed model in detail.
Section~\ref{sec:experiments} presents the experimental validation.
{ Section~\ref{sec:analysis} presents the comprehensive analysis on the effectiveness of the proposed model.}
Finally, Section~\ref{sec:conclusion} concludes the paper.

\section{Related Work and Background}
\label{sec:relate}
\subsection{Related Saliency Models}

\textit{Distinctiveness-Based Models}.
In distinctiveness-based models, the image regions with distinctive features in the local surroundings or global extent are regarded as salient regions. Itti \etal \cite{Itti1998} proposed a representative work of distinctiveness-based models. Specifically, they decomposed an image into a set of feature channels, i.e., the colors, intensity, and orientations of the luminance. Then, the center-surround local absolute difference was used to extract the local distinctiveness map in each feature channel. Finally, the distinctiveness maps of all the channels were integrated as the saliency map. This framework inspired many saliency models. For example, Fang \etal \cite{fang2016learning} computed the flexible center-surround difference in the principal-component-based subspaces of the image. Ishikura \etal \cite{ishikura2018extreme} computed the local extreme values in the CIELab color space.
Instead of the local distinctiveness,
Harel \etal \cite{Harel2006} proposed the use of a graph to globally interrelate the image regions.
The spatially weighted absolute difference of features between image regions was set as the edge between nodes in the graph.
Then, the global distinctive features were highlighted through a random walker process.
Goferman \etal \cite{Goferman2012a}, and Duan \etal \cite{duan2011visual}, computed a similar spatially weighted difference between image patches.
However, unlike that the work of Harel \etal \cite{Harel2006}, the patch's differences to all the other patches are simply summed to represent the global distinctiveness.
Further, Goferman \etal \cite{Goferman2012a} computed the difference in the CIElab color space, while Duan \etal \cite{duan2011visual} computed the difference in the dimension-reduced principal component space.
{
Based on similar scheme, Zhou and Jin \cite{zhou2013new} proposed a multi-scale operation to obtain the final saliency map.}

In addition to computing the difference of features,
the probability-based approach is a popular method to measure the distinctiveness of features.
Principal component analysis (PCA) and independent component analysis (ICA) are widely used to extract structure-related components according to image patches.
Based on the statistics of the PCA or ICA coefficients on the objective image (e.g., Refs.~\citenum{Torralba2003modeling,torralba2006contextual,imamoglu2013saliency}), or on a set of natural images (e.g., Refs.~\citenum{zhang2008sun,Bruce2006saliency,Borji2012}),
the detection of the distinctive features becomes a concept of rarity.
Specifically, a patch is more likely to be salient if it has rare PCA- or ICA-based features.

{
\textit{Sparsity-Based Models}
By assuming that an image can be represented in a redundant part and a sparse salient part \cite{Yan2010,lang2012saliency,wang2018visual}, 
many sparsity-based models are based on the low-rank matrix recovery theory \cite{candes2011robust}.
Particularly, Wang \etal \cite{wang2018visual} made an in-depth analysis of the sparse codes and the sparse reconstruction error that yields an effective saliency model. Besides, similar to the distinctiveness-based model, Rigas \etal \cite{rigas2015efficient} computed the saliency map via measuring the global distinctiveness of patches based on Hamming distance of sparse coding coefficients.

}

\textit{Spectral Models}.
The first spectral saliency model was proposed by Hou \etal \cite{hou2007saliency}, who suggested using the spectral residual of the amplitude spectrum to obtain the saliency map.
Later, Guo \etal \cite{guo2008spatio} suggested that the phase spectrum is the key for saliency detection.
Li \etal \cite{li2013visual} proposed a hypercomplex Fourier transform with a multiscale strategy for saliency detection.
Li \etal \cite{li2015finding} developed a learning-based filter with respect to phase spectra
to minimize the difference between the phase spectrum of the input image and fixation density map.

\textit{Learning-Based Models}.
The learning-based detectors of the semantic cues, e.g., face, human, and car, are proved to be beneficial for saliency detection \cite{Judd2009Learning,borji2012boosting}.
Moreover, benefiting from the powerful deep neural network, many recent saliency models achieved excellent performance in fixation prediction, e.g., Refs.~\citenum{vig2014edn,liu2015MrCNN,kruthiventi2017deepfix,liu2018deep,wang2018deep,he2018catches}.

\textit{The Post-Processing}.
Beyond the saliency-detection algorithms, post-processing could further enhance the performance of saliency detection.
For example, the fixation is more likely to be located at the image center \cite{tatler2007central}.
Thus, the border cut and center post-weighting that can highlight the center of the image will benefit saliency detection \cite{zhang2008sun,tatler2005visual,duan2011visual,Goferman2012a}.
However, the center regions will not always be salient.
Moreover, many saliency models would over-highlight the boundaries of the salient regions.
The Gaussian post-blurring, as described in Refs.~\citenum{Itti1998,Harel2006,fang2016learning}, can then extend the large values from the boundaries to the inner regions.
However, it will simultaneously highlight the unsalient exteriors.

{
\textit{Beyond Static Images.}
Saliency detection in the static images is the focus of this study.
In addition, saliency detection in videos and 360{\textdegree} (omnidirectional) images/videos has attracted significant research interests. In particular,
viewers usually do not have a complete view of the 360{\textdegree} content. Thus, the corresponding head rotation poses a problem for saliency detection \cite{sitzmann2018saliency}.
Moreover, when shown in equirectangular projection, the 360{\textdegree} contents are non-Euclidean \cite{bogdanova2008visual}.
To address these problems, many studies on 360{\textdegree} images/videos in viewing databases \cite{rai2017dataset,david2018dataset}, image quality assessment \cite{rai2017saliency}, and saliency detection \cite{de2017look}, have been conducted in recent years. 

}


\subsection{Image Quality Assessment}
\begin{figure}
    \centering
    \includegraphics[width=0.8\linewidth]{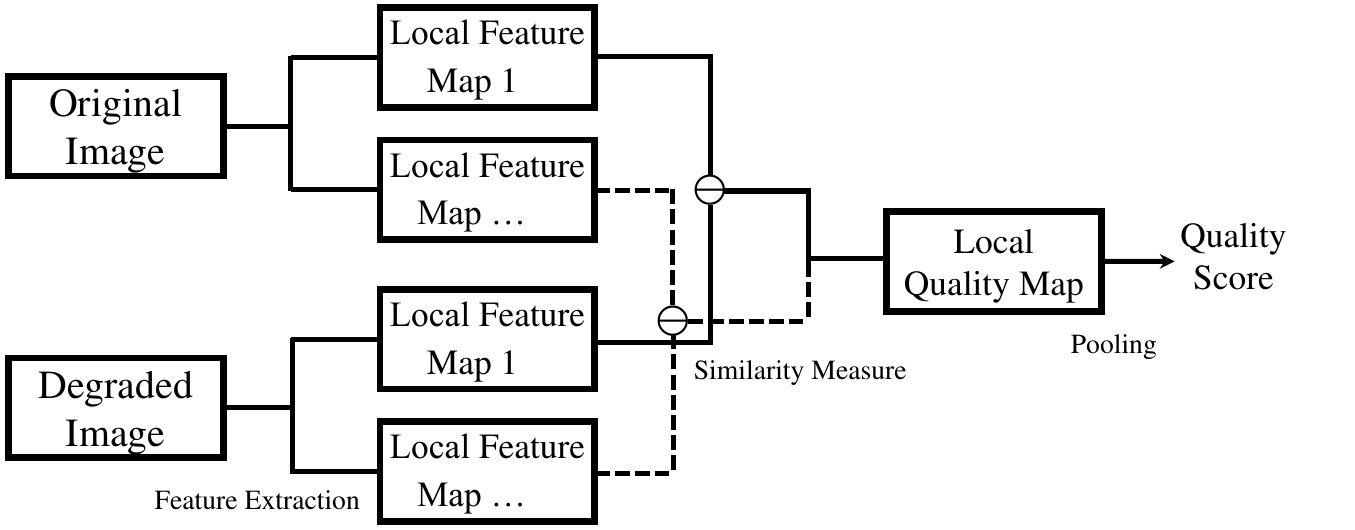}
    \caption{Typical framework of the structural-similarity-based FR-IQA models.}
    \label{fig:iqaframe}
\end{figure}

To bridge the knowledge gap, a brief introduction of IQA is presented.
Based on the availability of the pristine image, the IQA models can be classified into FR models, reduced reference models, and no reference models.
Our study focuses on the FR models
because we use this type of model to measure the structural similarity between two patches.
\figref{fig:iqaframe} shows a typical framework of the structural-similarity-based FR-IQA models.
Generally, the IQA models decompose the images into various local-feature maps.
As in many saliency models, the features in the IQA models are typically common in computer vision, for example, the features that could capture the local structures (i.e., the structural features) and the chrominance features.
The point-by-point similarities in different maps can be computed and combined to obtain a local quality map.
Then, a pooling step, such as the mean average or standard deviation, is executed on the local quality map to obtain the overall quality score of the degraded image.

Many IQA models follow this framework.
For example, the well-known SSIM \cite{wang2004image}
extracts three features of images: local mean, LC, and local structure.
Then, the point-wise similarity comparisons of the luminance, contrast, and
structures are computed, combined, and pooled to yield the IQA model.
The main motivation to use the structural similarity is that the HVS is highly sensitive to the local structure.
This idea inspired many IQA models.
Xue \etal \cite{Xue2014} proposed the use of the GM as the only structural feature for measuring structural similarity.
Specifically, the GM is the root mean square of image directional gradient along the horizontal and vertical directions.
Some works {\color{blue}(see~\citenum{mou2014,Xue_2014})} validated that the Laplacian-of-Gaussian (LOG) signal
acts as an effective and efficient structural feature for IQA.
Moreover, the feature similarity metric \cite{zhang2011fsim} used the phase congruency and GM of the \textit{Y} component and the pixel value of \textit{IQ} chrominance components in the \textit{YIQ} color space.
The saliency map is used in the visual saliency-induced index \cite{Zhang2014} as a feature map, together with the GM of \textit{L} component, and the pixel value of the \textit{MN} chrominance components in the \textit{LMN} color space.

To measure the similarity between the feature maps,
the introduced IQA models mainly used a correlation-based measure as follows:
\begin{equation}
\text{sim} = \dfrac{2 \bm{f}(p)\cdot \bm{r}(p) + c}{\bm{f}(p)^2+\bm{r}(p)^2 +c},
\label{equ:corr}
\end{equation}
where $\bm{f}$ and $\bm{r}$ are the feature maps of the original and degraded image, respectively, $p$ denotes the point-by-point computation for pixel $p$, and $c$ is a small positive constant.

The above similarity measure can simply be deduced as a dissimilarity measure, i.e.,
\begin{equation}
\label{equ:ssimDis}
\bm{d}(p) = 1-\text{sim}= \frac{(\bm{f}(p)-\bm{r}(p))^2}{\bm{f}(p)^2+\bm{r}(p)^2+c}.
\end{equation}
This dissimilarity measure can be regarded as the normalized version of the squared absolute difference, i.e.,
\begin{equation}
\label{equ:mse}
\bm{d}(p) = |\bm{f}(p)-\bm{r}(p)|^2.
\end{equation}
The denominator $\bm{f}(p)^2+\bm{r}(p)^2+c$ makes \equref{equ:ssimDis} a relative dissimilarity measure, which is claimed to be consistent with Weber's law in Ref.~\citenum{wang2004image}.
In IQA models, parameter $c$ is used to avoid division by zeros \cite{wang2004image}.
In this study, we show that parameter $c$ serves as a normalization parameter to control the effects of the insignificant features in the proposed model.
Only when the energy of the comparing features (i.e., $\bm{f}(p)^2+\bm{r}(p)^2$) approaches or exceeds the level of $c$, can their absolute difference affect the result computed by \equref{equ:ssimDis} (for details, refer to \secref{sec:analysis}).
The comprehensive analysis of the advantage of the correlation-based dissimilarity measure can be found in Refs.~\citenum{Seshadrinathan2008,Brunet2012}.

\section{Method}

\begin{figure}
    \centering
    \includegraphics[width=0.95\linewidth]{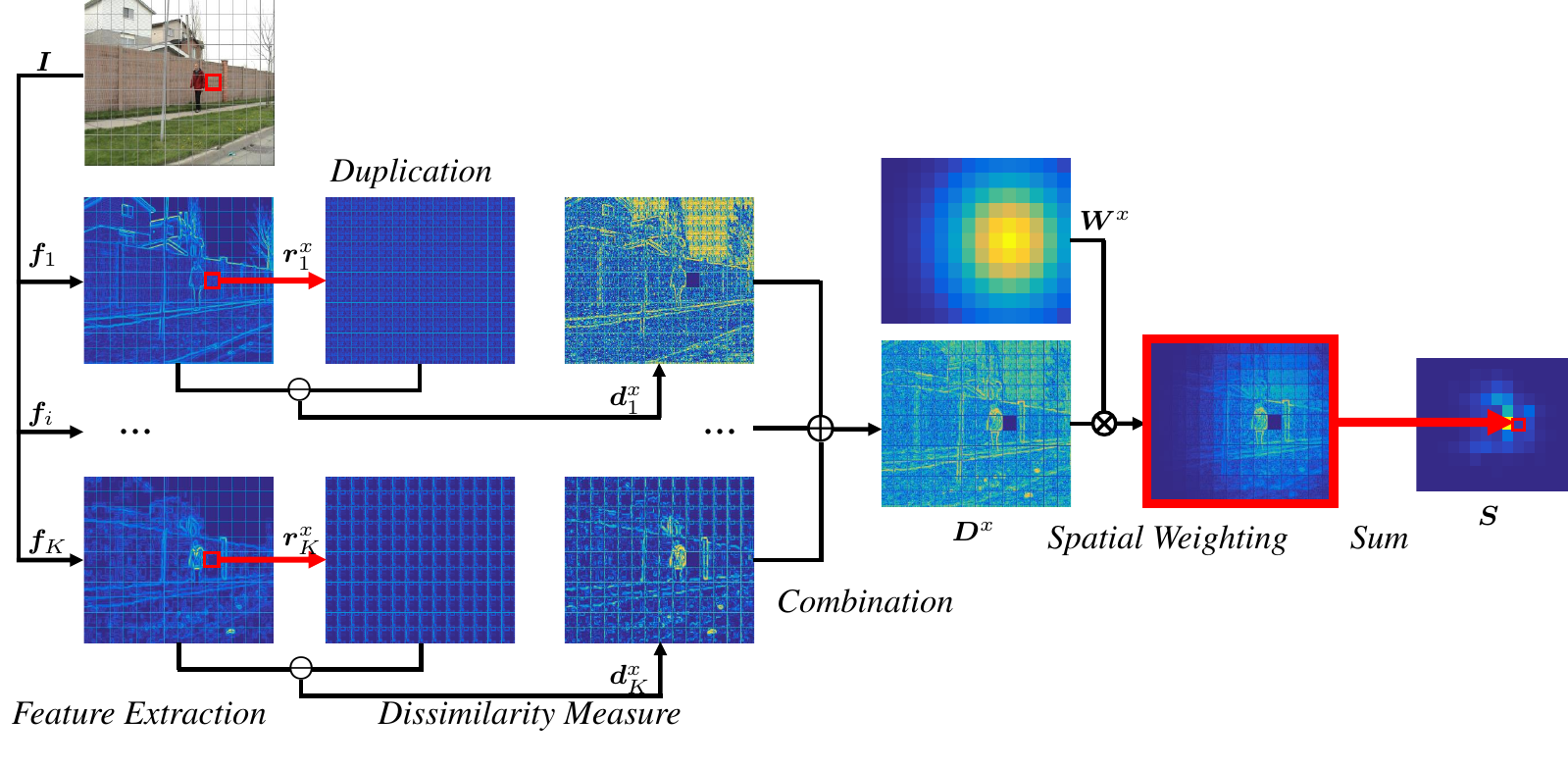}
    \caption{Proposed framework to compute the saliency of the patch marked with the red box. The selected patch is compared with other patches through the structural dissimilarity measure induced by the IQA model. Sample feature maps $\bm{f}_i$ are the GM in the \textit{YIQ} color space.}
    \label{fig:framework}
\end{figure}

In this paper, we propose the use of structural dissimilarity induced by IQA models as the distinctiveness measure for saliency detection.
To consider computation complexity, the comparison between patches was conducted in a nonoverlapped manner.
To facilitate nonoverlapping of the split,
we resized the image to multiples of patches $[Mw\times Nw]$,
where $w$ is the width of the patch in pixels, $M$ and $N$ are the numbers of patches in height and width of the image.
We experimentally set $w=24$, and $\mathrm{min}(M,N)=11$.
\figref{fig:framework} illustrates the process to compute the saliency of patch $x$, which is marked with a red box on the resized image $\bm{I}$.
The process involves four main stages.
\label{sec:methods}

%

1) \textit{Feature Extraction}.
To compute the structural dissimilarity between image patches,
$\bm{I}$ is first decomposed into a set of structural features.
According to studies, such as Refs~\citenum{ishikura2018extreme,van2006boosting,Itti1998},
both the luminance and chrominance information are crucial for saliency detection.
Therefore, structural features are proposed to be extracted in each color channel of the \textit{YIQ} color space.
Suppose there are $K$ channels,
then the feature maps are indicated by $\bm{f}_i$ ($i=1,2,...,K$).

2) \textit{Dissimilarity Measure}.
The correlation deduced relative difference measure (i.e., \equref{equ:ssimDis}) is proposed to compute the point-wise dissimilarity between feature patches.
In each feature channel, we duplicated patch $x$ in a nonoverlapped manner to obtain duplication feature map $\bm{r}_i^x$, as shown in \figref{fig:framework}.
Then, the global comparison of $x$ with all other patches is equivalent to the comparison between $\bm{r}_i^x$ and $\bm{f}_i$.
We rewrite the computation of dissimilarity map $\bm{d}_i^x$ as
\begin{equation}
\bm{d}_i^x(p) = \frac{(\bm{f}_i(p)-\bm{r}_i(p))^2}{\bm{f}_i(p)^2+\bm{r}_i(p)^2+c}.
\label{equ:newD}
\end{equation}

3) \textit{Combination}. The dissimilarity maps in different feature channels are combined to an overall dissimilarity map $\bm{D}^x$.
Specifically,
\begin{equation}
\bm{D}^x(p) = \left(\sum_{i=1}^K {\bm{d}_i^x(p)} \right)^\frac{1}{2}.
\label{equ:disD}
\end{equation}

4) \textit{Spatial Weighting and Sum}. Spatially weighted $\bm{D}^x$ are summed to predict the saliency of $x$.
This spatial weighting has been adopted in many works (e.g., Refs~\citenum{Harel2006,Goferman2012a,duan2011visual}) and
 controls the extent of the global comparison.
Specifically, spatial weighting map $\bm{W}^x$ is the distance of patch $x$ to other patches.
Let $\bold{x}$ and $\bold{y}$ indicate the location of patch $x$ and $y$, respectively.
The spatial distance between $x$ and $y$ is computed by
$\textrm{exp}(-\frac{\|\bold{x}-\bold{y}\|^2}{2\sigma^2})$,
where $\sigma^2$ is a scale parameter.
To ensure the same size for $\bm{W}^x$ and $\bm{D}^x$, the points in $y$ share the same distance value.
Finally, the spatially weighted global dissimilarity of $x$ (i.e., the saliency of $x$) is summed as
\begin{equation}
\bm{S}(x)=\left(\sum_{p\in{\bm{D}^x}}\left(\bm{D}^x(p)\cdot \bm{W}^x(p)\right)\right)^\theta.
\label{equ:saliencyS}
\end{equation}

The global sum of the relative differences would result in a slight difference between the saliency of patches.
Therefore, this paper proposes the exponential operation by using exponent $\theta$ to highlight the salient patches. We empirically set $\theta=8$.
It is worth noting that the relative values in the saliency map will not be changed.

The abovementioned process was repeated until the saliency of all the patches is obtained.
Then, the saliency map was normalized to the dynamic range of $[0,1]$.
No other post-processing operation, e.g., Gaussian post-blurring or center post-weighting, was conducted.

\section{Experiments}
\label{sec:experiments}

\subsection{Experimental Protocols}

\subsubsection{Features}
The structural dissimilarity based on the correlation of the structural features is crucial in the proposed framework.
To validate the effectiveness of the proposed framework, we adopted two widely used structural features: LC and GM.
Let the structural-dissimilarity-based saliency (SDS) denote the proposed framework. The two corresponding models are then denoted as $\text{SDS}_\text{GM}$ and $\text{SDS}_\text{GM}$, respectively.

\subsubsection{Databases}
The performance of the saliency models depends on the consistency between the saliency map and the corresponding eye fixations.
Three popular saliency databases were used in this study: {Toronto} \cite{Bruce2006saliency}, {MIT} \cite{Judd2009Learning} and {ImgSal} \cite{li2013visual}.
The {Toronto} database consists of 120 images with human fixation data recorded on 20 viewers,
while the {MIT} database consists of 1003 images and the corresponding eye fixation data was recorded on 15 viewers.
Both {Toronto} and {MIT} databases have natural indoor and outdoor scenes.
However, {MIT} contains many daily-life and portrait pictures, making it a more challenging database.
The {ImgSal} database consists of 235 images and the corresponding eye fixation data was recorded on 21 viewers.
The images in {ImgSal} were classified into six categories:
50/80/60 images with large/intermediate/small salient regions, respectively, 15 images with a cluttered background, 15 images with repeating distractors, and 15 images with both large and small salient regions.

\subsubsection{Evaluation metrics}
Many evaluation metrics exist for the evaluation of saliency prediction performance.
According to the comprehensive study on the evaluation metrics by Bylinskii \etal \cite{bylinskii2018different},
the normalized scanpath saliency (NSS) and the linear correlation coefficient (CC) are the most reasonable metrics, and are recommended to be selected.

The NSS computes the mean value at the fixation locations on a z-score-normalized saliency map.
Therefore, the larger NSS scores are desirable.
Let $\mathrm{SM}$ denotes the saliency map of a given image, and $\mathrm{FM}$ denotes the corresponding fixation map (i.e., the eye fixation points of the image).
The NSS score can be computed as
\begin{equation}
\label{equ:NSS}
\begin{split}
\mathrm{NSS} &= \frac{1}{\sum_p{bool(\mathrm{FM}_p})} \sum_p {(\mathrm{\overline{SM}}_p \times{bool(\mathrm{FM}_p}))},\\
&\text{where } \mathrm{\overline{SM}} = \frac{\mathrm{SM}-\mu{(\mathrm{SM})}}{\sigma{(\mathrm{SM}})},\\
&bool(\mathrm{FM}_p)=
\begin{cases}
1,& \text{$\mathrm{FM}_p>0$}\\
0,& \text{$\mathrm{FM}_p=0$}
\end{cases}
,
\end{split}
\end{equation}
where $\mu$ and $\sigma$ compute the mean and standard deviation of the map, respectively, and $p$ denotes the point-wise operation.

The CC computes the Pearson's linear correlation coefficient between the saliency map and Gaussian-blurred fixation map (i.e., the fixation density map).
We still use $\mathrm{FM}$ to denote the Gaussian-blurred fixation map.
The CC score can be computed as
\begin{equation}
\label{equ:CC}
\text{CC} = \frac{cov(\mathrm{SM},\mathrm{FM})}{\sigma{(\mathrm{SM})}\times \sigma{(\mathrm{FM})}},
\end{equation}
where $cov$ indicates the covariance computation.
The CC has a value in the range of $[-1, +1]$, where $\text{CC}=1$ is the total positive linear correlation.

In addition to the NSS and CC scores, in the following experiments, we report on the scores of two historically widely used metrics: the earth mover's distance (EMD) and area under the ROC curve (AUC) \cite{Judd2009Learning}.
A lower EMD and higher AUC indicate a better consistency between the saliency and fixation maps.
In the four evaluation metrics, the recommended NSS has the lowest computation complexity.
Therefore, in the following analysis of the effects of the parameters and dissimilarity measure, we only report on the NSS scores.
Then, in the comprehensive comparison with other competing saliency models, we report the scores evaluated by all the four metrics.
Specifically, we adopted the codes of the four evaluation metrics provided by Ref.~\citenum{mit-saliency-benchmark}.

\subsubsection{Competing saliency models}
To demonstrate the performance of the proposed SDS, we compared it with 11 state-of-the-art and representative saliency models, including IT \cite{Itti1998}, GBVS \cite{Harel2006}, JUDD \cite{Judd2009Learning}, SWD \cite{duan2011visual}, CA \cite{Goferman2012a}, FES \cite{tavakoli2011fast}, HFT \cite{li2013visual}, SSD \cite{li2015finding}, BMS \cite{zhang2016exploiting}, and LDS \cite{fang2016learning}.
All the source codes are publicly available on the Internet. We used the default parameters provided in the source codes for all models.
In these models, HFT and SSD are the spectral saliency models, JUDD and LDS are the learning-based models, and the other models are distinctiveness-based models.
Moreover, IT, GBVS, FES, HFT, BMS, and LDS adopt the Gaussian post-blurring, whereas
JUDD, SWD, and CA adopt the center post-weighting.

The saliency maps generated by the various saliency models differed in size.
Thus, a resizing process was embodied in the codes of the evaluation metrics provided by Ref.~\citenum{mit-saliency-benchmark}.
All the saliency maps were cubic-interpolated to the same size as the fixation map.
In addition, all the saliency maps were normalized to the dynamic range of $[0,1]$ before evaluation.

\subsection{Effects of the Parameters}
\label{sec:effectsofc}
\begin{figure}
    \centering
    \includegraphics[width=1.0\linewidth]{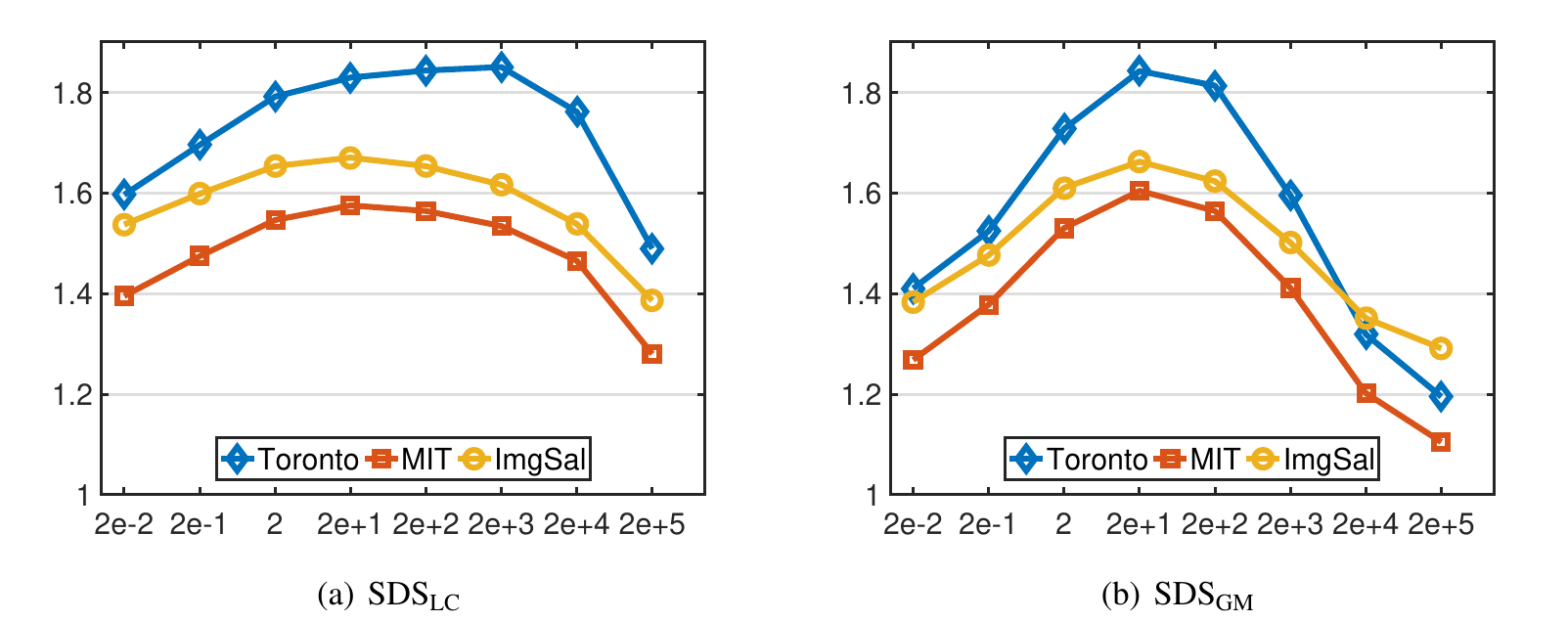}
    \caption{NSS scores of $\text{SDS}_\text{LC}$ and $\text{SDS}_\text{GM}$ against $c$ over the three databases.}
    \label{fig:3setC}
\end{figure}

There are two important parameters in the proposed model: normalization parameter $c$ in the structural dissimilarity computation (i.e., \equref{equ:disD}) and scale parameter $\sigma^2$ in the spatial weighting computation (i.e., \equref{equ:saliencyS}).
Normalization parameter $c$ controls the effects of the insignificant features.
For comprehensive analysis, we report the saliency detection performance with respect to parameter $c$ over the three adopted databases of $\text{SDS}_\text{LC}$ and $\text{SDS}_\text{GM}$ in terms of the NSS score.
The results are shown as scatter plots in \figref{fig:3setC}.
As shown, considering the wide range of $c$, both $\text{SDS}_\text{LC}$ and $\text{SDS}_\text{GM}$ are not very sensitive to the changing of $c$.
However, $\text{SDS}_\text{LC}$ is more insensitive
 because feature LC usually possess a larger value than GM.
In addition, the best NSS scores of $\text{SDS}_\text{GM}$ on the three databases are achieved when $c=20$.
Similar results can be observed for $\text{SDS}_\text{LC}$, except on the Toronto database, for which the best NSS score is achieved when $c=2000$.
Considering the overall performance on the three databases, we set $c=20$ for both $\text{SDS}_\text{LC}$ and $\text{SDS}_\text{GM}$ in the following experiments.


\begin{figure}
    \centering
    \includegraphics[width=1.0\linewidth]{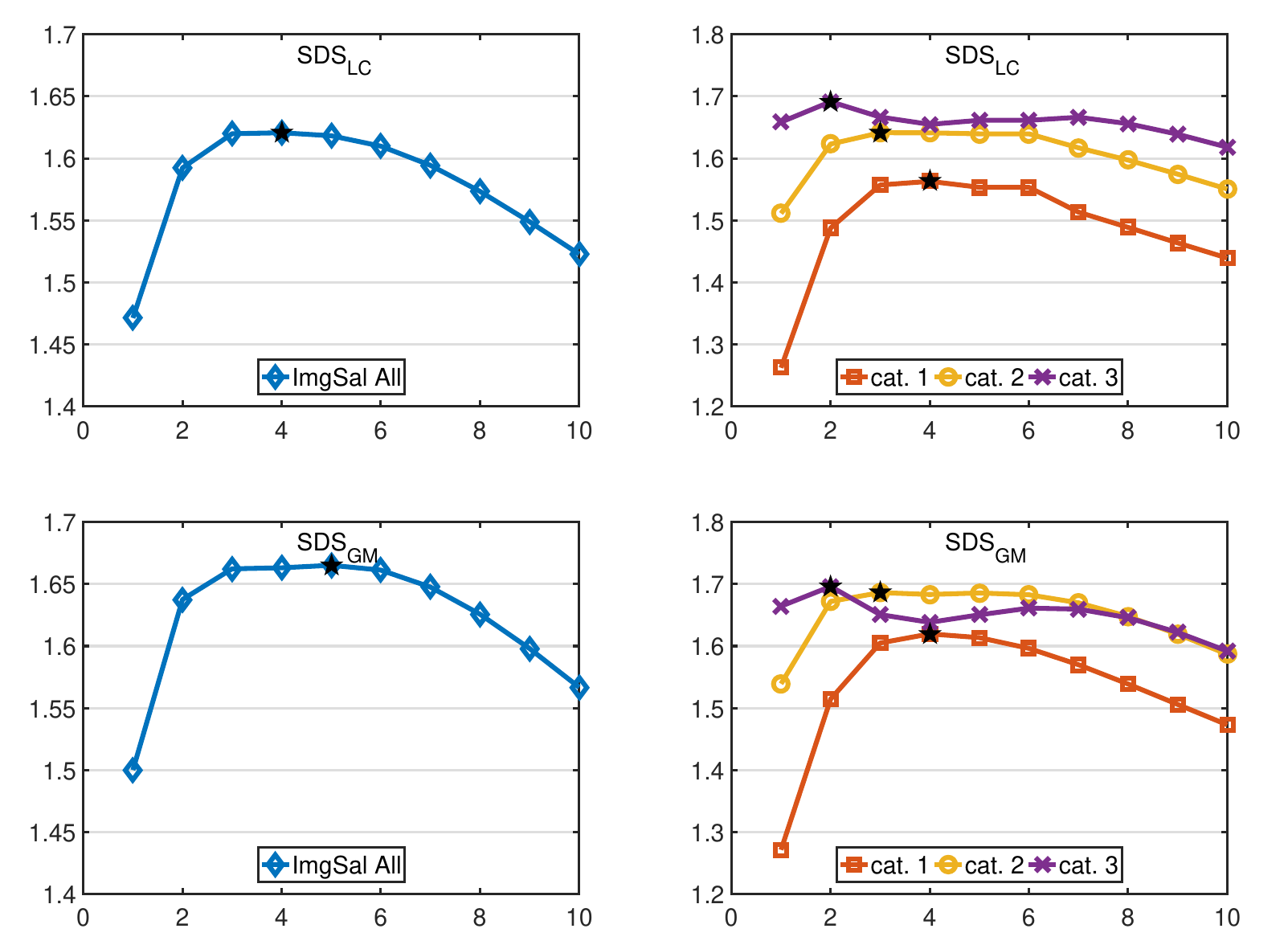}
    \caption{NSS scores of $\text{SDS}_\text{LC}$ and $\text{SDS}_\text{GM}$ against $\sigma^2$. The value of $\sigma^2$ varies from 1 to 10. The left subfigure shows the NSS scores over the whole {ImgSal} database. The right subfigure shows the NSS scores of the six categories. The maximum values are marked by the pentagram.}
    \label{fig:imgsalplot}
\end{figure}

Scale parameter $\sigma^2$ in the spatial weighting computation controls the extent of global comparison, which in turn
 affects the ability to detect various sized salient regions.
Taking advantage of the {ImgSal} database containing categories with various sized salient regions, we analyzed the detection performance of $\text{SDS}_\text{LC}$ and $\text{SDS}_\text{GM}$ under different $\sigma^2$.
The specific NSS scores against $\sigma^2$ are shown in \figref{fig:imgsalplot}.
In particular, we also report the detection performances on categories 1, 2, and 3, in which the sizes of salient regions showed a decreasing trend.


\begin{figure}
    \centering
    \includegraphics[width=1.0\linewidth]{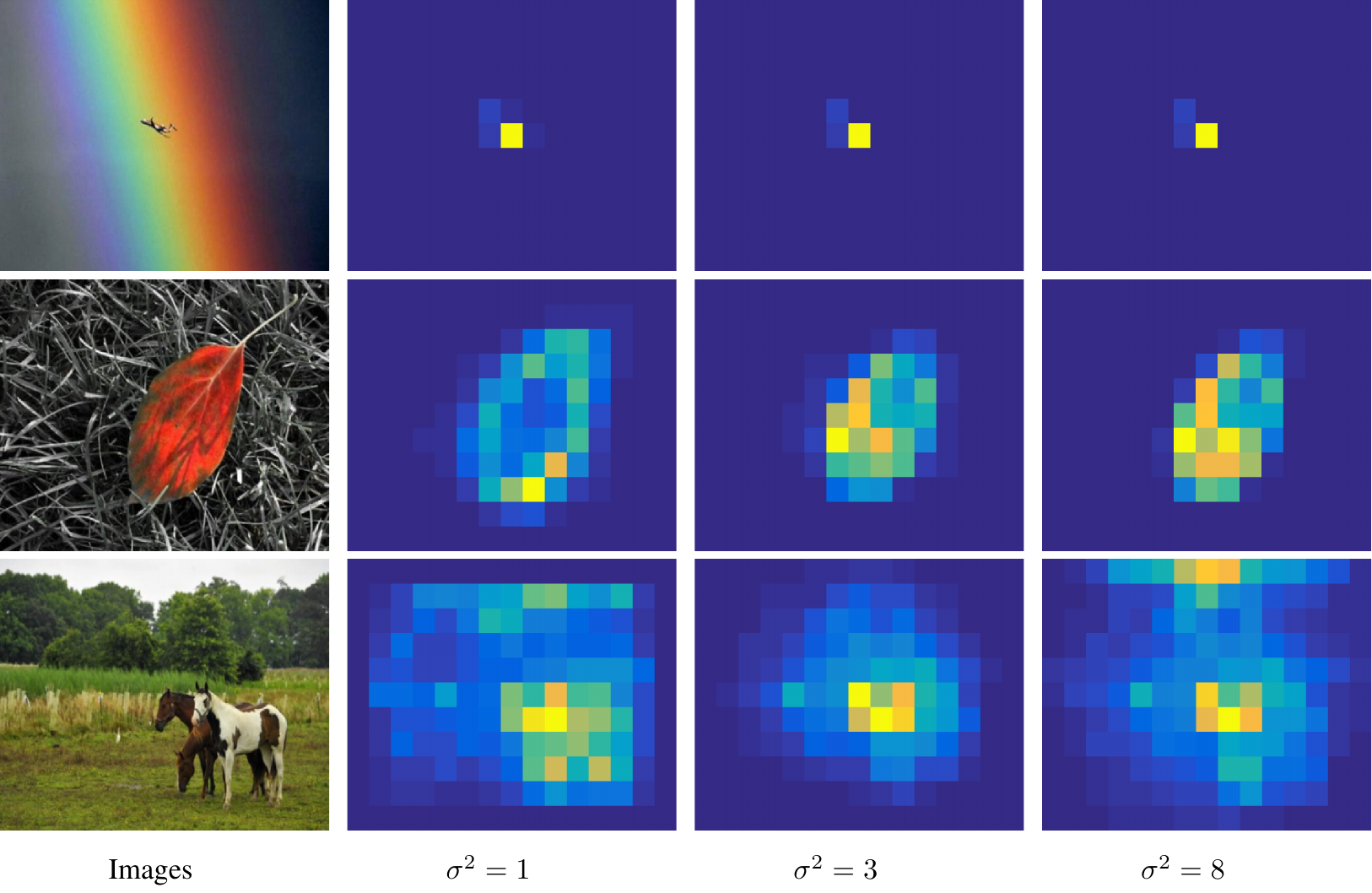}


    \caption{Illustration of the effects of $\sigma^2$ for various sized salient regions. The first column shows three example images. The remaining are the enlarged saliency maps generated by $\text{SDS}_\text{GM}$ under various $\sigma^2$.}
    \label{fig:sigmaillu}
\end{figure}

\figref{fig:imgsalplot} shows that the NSS scores are almost the same when $\sigma^2=3\text{ to }6$ on the whole database.
By contrast, apparent changes can be observed at both small and large $\sigma^2$.
In addition, the NSS scores for categories~1 and 2 (i.e., large and intermediate salient regions, respectively) are greatly increased at the beginning of the increase of $\sigma^2$.
However, the NSS scores for category~3 (i.e., small salient regions) are almost the same at various $\sigma^2$.
This is because the small salient region will be dissimilar to both the local and the global surroundings, as shown in the first row of \figref{fig:sigmaillu}.
For a large salient region, the dissimilarity measure in the local extent will tend to highlight the boundaries of the region.
By contrast, its inner region, especially the inner homogeneous region, will be highlighted by a global dissimilarity measure, as shown in the second row of \figref{fig:sigmaillu}.
However, a slight performance drop can also be observed at the large $\sigma^2$.
This is because the unsalient regions might be incorrectly highlighted by the global dissimilarity measure, as illustrated in
 the last row of \figref{fig:sigmaillu}.
When measured in the global extent, the patches at the sky, which are distinct to the vast grass and trees, are highlighted on a large $\sigma^2$.

Considering the scatter plots in \figref{fig:imgsalplot},
the best NSS scores are reached at $\sigma^2=4,3,2$ for categories~1,2, and 3, respectively, both for $\text{SDS}_\text{LC}$ and $\text{SDS}_\text{GM}$.
Therefore, considering the overall performance on different sized salient regions, we fixed $\sigma^2=3$ for both $\text{SDS}_\text{LC}$ and $\text{SDS}_\text{GM}$ in the following experiments.

\subsection{Comparison with the State-of-the-art models}

The mean scores evaluated by all four metrics on the three databases were calculated to compare the proposed $\text{SDS}_\text{LC}$ and $\text{SDS}_\text{GM}$ with other competing saliency models.
\tabref{tab:allperformance} summarizes the results.

\begin{table*}
\caption{Performance comparison of the proposed $\text{SDS}_\text{LC}$, $\text{SDS}_\text{GM}$ with other competing saliency models. The best and
 the second-best results are marked by red and blue colors, respectively.}
\label{tab:allperformance}
\centering
\scriptsize
\begin{tabular}{lcccccccccccccc}
\toprule

&\multicolumn{4}{c}{{Toronto} (120 images)} &\multicolumn{4}{c}{{MIT} (1003 images)} &\multicolumn{4}{c}{{ImgSal} (235 images)} &\multicolumn{2}{c}{\textit{Average}}\\
\cmidrule(lr){2-5}
\cmidrule(lr){6-9}
\cmidrule(lr){10-13}
\cmidrule(lr){14-15}
& EMD & AUC & CC & NSS &EMD & AUC & CC & NSS &EMD & AUC & CC & NSS  & CC & NSS\\
\toprule
IT \cite{Itti1998}              &2.837 &0.799 &0.503 &1.297 &5.146 & 0.767 & 0.334 & 1.093&2.043&	0.799&	0.577&	1.325 &  0.391 & 1.151\\
GBVS \cite{Harel2006}           &2.237	&0.830	&0.600	&1.519 & 4.310 & 0.821 & 0.421 & 1.366&1.777&	\color{blue}\textbf{0.833}&	0.674&	1.564 & 0.480 & 1.414\\
JUDD \cite{Judd2009Learning}    &3.037 &0.839 &0.563&1.389&5.329 & 0.835 &0.419 & 1.329 & 2.706 & \color{red}\textbf{0.837} & 0.628 & 1.429 &0.468 &1.352\\
LG \cite{Borji2012}             & 3.443 & 0.753 & 0.386 & 1.072 & 5.764 & 0.746 & 0.292 & 0.987&3.213&	0.696&	0.373&	0.883 &  0.314 & 0.977\\
CA \cite{Goferman2012a}         & 3.041 & 0.781 & 0.465 & 1.272 &5.304&0.755&0.313&1.052&2.374&0.791&0.577&1.384&0.353&1.110\\
SWD \cite{duan2011visual}       & 2.564 & 0.836 & 0.608 & 1.524 & 4.692 & 0.831 & 0.446 & 1.438&2.312&	0.830&	0.648&	1.485 & 0.495 & 1.454\\
FES \cite{tavakoli2011fast}     & 2.132 & 0.804 & 0.593 & 1.569 & 3.700 & 0.807 & 0.450 & 1.476&1.888	&0.787	&0.625	&1.513 & 0.493 & 1.491\\
HFT \cite{li2013visual}         &2.283 & 0.825& 0.606 &1.631 &4.304 &0.809 &0.422&1.395&1.709&0.821&{0.695}&\color{blue}\textbf{1.670}&0.486&1.463\\
SSD \cite{li2015finding}        & 2.222 & 0.805 & 0.590 & 1.645 & 4.104 & 0.800 & 0.433 & 1.450&1.957	&0.792	&0.589	&1.431 & 0.474 & 1.464\\
BMS \cite{zhang2016exploiting}  & 3.115 & 0.786 & 0.518 & 1.560 & 5.348 & 0.785 & 0.374 & 1.314&2.625	&0.788	&0.620	&1.599 &0.429 & 1.385\\
LDS \cite{fang2016learning}     & 1.825 & \color{red}\textbf{0.845} & 0.647 & 1.756 & 3.554 & \color{blue}\textbf{0.838} & 0.483 & \color{blue}\textbf{1.582}&1.634	&0.828	&0.694	&1.651 & 0.534 & 1.609\\
\cmidrule(lr){1-1}
\cmidrule(lr){2-5}
\cmidrule(lr){6-9}
\cmidrule(lr){10-13}
\cmidrule(lr){14-15}

$\text{SDS}_\text{LC}$          & \color{blue}\textbf{1.753} & 0.841 & \color{blue}\textbf{0.688} & \color{blue}\textbf{1.829} & \color{red}\textbf{3.466} & 0.838 & \color{blue}\textbf{0.486} & 1.575 & \color{blue}\textbf{1.485} & 0.830 &\color{red}\textbf{0.706}&\color{red}\textbf{1.670}  & \color{blue}\textbf{0.542} & \color{blue}\textbf{1.614}\\
$\text{SDS}_\text{GM}$          & \color{red}\textbf{1.735} & \color{blue}\textbf{0.844} & \color{red}\textbf{0.694} & \color{red}\textbf{1.842} & \color{blue}\textbf{3.471} & \color{red}\textbf{0.841} & \color{red}\textbf{0.493} & \color{red}\textbf{1.604}&\color{red}\textbf{1.483} & {0.831}&\color{blue}\textbf{0.702}&1.662  & \color{red}\textbf{0.547} & \color{red}\textbf{1.635}\\
\bottomrule
\end{tabular}

\end{table*}

The scores of the proposed $\text{SDS}_\text{GM}$ and $\text{SDS}_\text{LC}$ can be observed to be the same on the three databases.
$\text{SDS}_\text{GM}$ outperforms all the other competing models with respect to the EMD, CC, and NSS scores on the Toronto and MIT databases.
$\text{SDS}_\text{GM}$ achieves the best CC and NSS scores on ImgSal.
Furthermore,
LDS achieves the best AUC score on Toronto and the second-best AUC and NSS scores on MIT.
JUDD and GBVS achieve the best and second-best AUC scores on ImgSal, respectively.
For better comparison of the performance, we calculated the weighted average of CC and NSS scores over the three databases.
The average weights are determined according to the number of images in the three databases.
\tabref{tab:allperformance} summarizes the results.
The table shows that $\text{SDS}_\text{GM}$, $\text{SDS}_\text{LC}$, and LDS achieve the top~3 positions, among which the $\text{SDS}_\text{GM}$ is the best.

\begin{figure}
    \centering
    \includegraphics[width=0.8\linewidth]{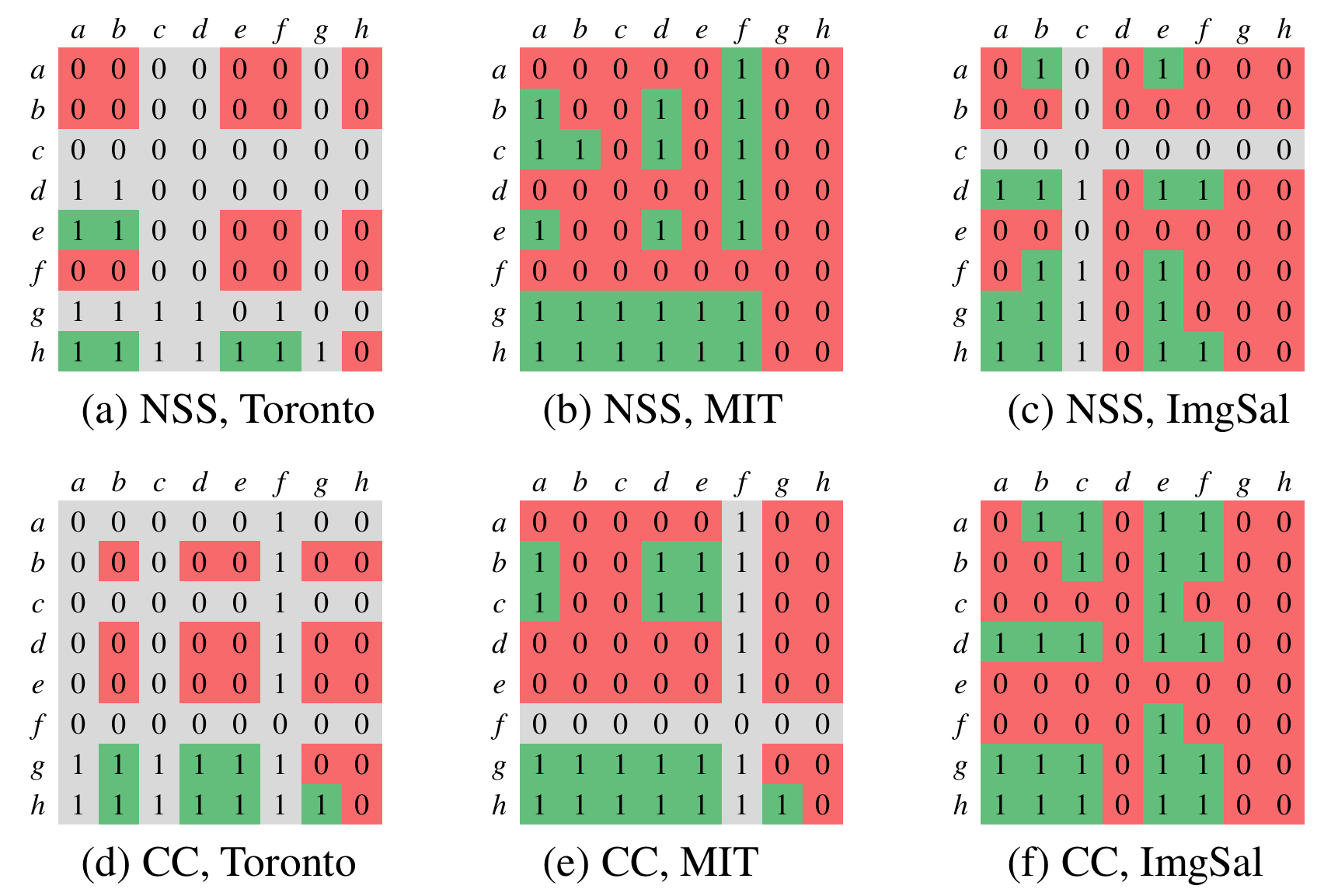}
    \caption{The results of statistical significance test of eight models. The eight models indicated by `\textit{a}'~to~`\textit{h}' are GBVS, SWD, FES, HFT, SSD, BMS, LDS, and the proposed $\text{SDS}_\text{GM}$. The captions of subfigures indicate the tested metrics and databases.
    A value of `0', colored in red, indicates no significant difference between two models, whereas
    a value of `1', colored in green, indicates the model in the row is significantly better than the model in the column. The models violating the normality distribution assumption are colored in gray. The proposed model performs significantly better than other models in most cases.
    }
    \label{fig:sig}
\end{figure}

{
For statistical significance test of mean scores in terms of NSS and CC,
We further conducted the t-test at $p\leq0.05$ level of significance \cite{borji2013quantitative}.
The results are summarized in \figref{fig:sig},
where a value of `1' indicates the model in row has a larger mean score than the model in the column and the two mean scores are statistically significantly different.
Moreover, we have checked the normality distribution assumption of the scores by the Jarque-Bera test \cite{jarque1980efficient}.
If a model violates this assumption, the corresponding results in \figref{fig:sig} are colored in gray.
Here, $\text{SDS}_\text{GM}$ is tested as the representative, because the proposed  $\text{SDS}_\text{GM}$ and $\text{SDS}_\text{LC}$ have similar performance.

It can be seen that on the Toronto database, the proposed model is significantly better than the other models in terms of both NSS and CC. Similar results can be observed on the MIT database in terms of CC. And the proposed model and LDS perform well with no significant difference in terms of NSS. On the ImgSal database, the proposed model and HFT perform well with no significant difference in terms of both NSS and CC.
Therefore, we can conclude that the proposed model is highly competitive to these advance models.
}


\begin{table}
\caption{NSS Scores for each category of the Toronto database. The numbers of images in all categories are shown in brackets. {The best and the second-best results are marked by red and blue colors, respectively.}}
\centering
\begin{tabular}{lccccccc}
\toprule
& cat. 1 & cat. 2 & cat. 3 & cat. 4 & cat. 5 &cat. 6 & all\\
& (50) & (80) & (60) & (15) & (15) & (15) & (235)\\
\toprule
IT & 1.241 & 1.300 & 1.460 & 1.141 & 1.213 & 1.494 & 1.325\\
GBVS & 1.406 & 1.553 & 1.648 & 1.538 & 0.790 & 1.618 & 1.564\\
JUDD & 1.354 & 1.432 & 1.434 & 1.353 & 1.707 & 1.454 &1.429\\
LG & 0.793 & 0.961 & 0.897 & 0.507 & 0.915 & 1.051 & 0.883\\
CA & 1.194 & 1.387 & 1.534 & 1.315 & 1.404 & 1.447 & 1.384\\
SWD & 1.498 & 1.466 & 1.487 & 1.295 & 1.712 & 1.495 & 1.485\\
FES & 1.541 & 1.590 & 1.249 & 1.486 & 1.975 & 1.629 & 1.513\\
HFT & 1.512 & 1.607 & \color{red}\textbf{1.727}& \color{red}\textbf{1.764} & \color{red}\textbf{2.114} & \color{red}\textbf{1.761} & \color{blue}\textbf{1.670}\\
SSD & 1.329 & 1.466 & 1.510 & 1.070 & 1.612 & 1.454 & 1.431\\
BMS & 1.597 & 1.542 & \color{blue}\textbf{1.680} & 1.436 & 1.732 & 1.620&1.599\\
LDS & \color{red}\textbf{1.621} & 1.617 & 1.634 & 1.509 & \color{blue}\textbf{2.078} & \color{blue}\textbf{1.695} &1.651\\
\midrule
$\text{SDS}_\text{LC}$          & 1.582 & \color{red}\textbf{1.702} & 1.659 & \color{blue}\textbf{1.554} & 1.994 & 1.634 &\color{red}\textbf{1.670}\\
$\text{SDS}_\text{GM}$          & \color{blue}\textbf{1.605} & \color{blue}\textbf{1.687} & 1.649 & 1.499 & 1.988 & 1.611&1.662\\

\bottomrule
\label{tab:imgsalcate}

\end{tabular}
\end{table}

\begin{figure}
    \centering
    \includegraphics[width=0.95\linewidth]{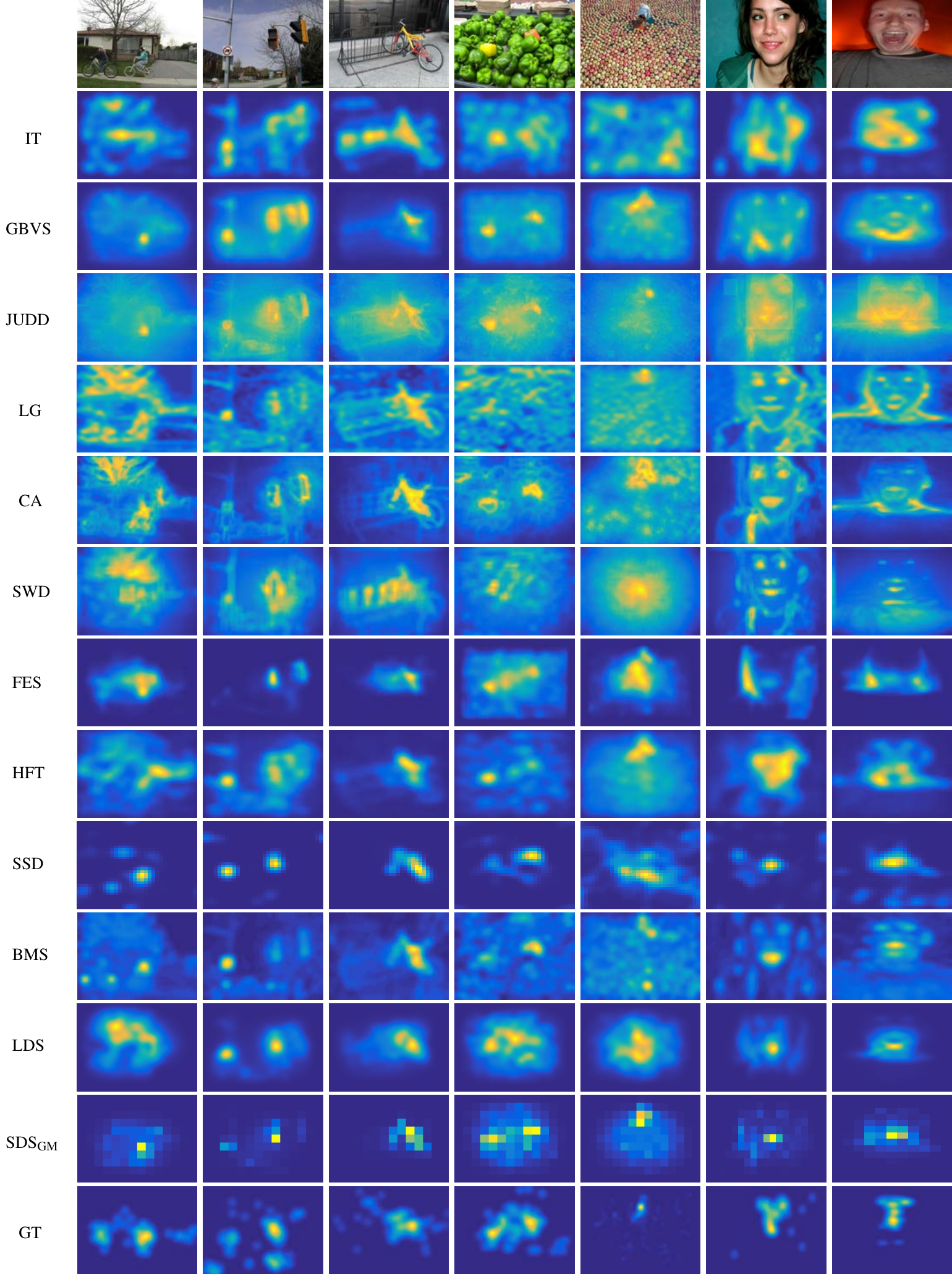}
    \caption{Representative saliency maps generated by the proposed {$\text{SDS}_\text{GM}$} and other 11 saliency models. The saliency maps of $\text{SDS}_\text{GM}$, in which the sizes of the short sides are all 11 pixels, are enlarged for better visual comparison. The GT indicates the ground-truth fixation density map.}
    \label{fig:visualcomp}
\end{figure}

To further validate the ability of the proposed models in handling the different-sized salient regions, we compared all the models over each category of {ImgSal}. \tabref{tab:imgsalcate} summarizes the corresponding NSS scores.
As shown, the HFT model achieves the best scores for categories~3--6.
Furthermore, LDS presents a promising performance in each category.
Note that the HFT model conducted a multiscale analysis in the spectral domain and designed a criterion to select the optimal scale that can best highlight the salient regions.
LDS conducted a flexible-scaled random contrast to highlight the distinctive features.
Therefore, the multiscale operation is a positive solution to handle different sized salient regions.
In contrast,
the proposed $\text{SDS}_\text{GM}$ and $\text{SDS}_\text{LC}$, without the multiscale operations, also present competitive results in all categories.
This effectiveness is due to the following two aspects.
First, the relative difference with normalization parameter $c$ can highlight the inner regions of the large salient regions.
Second, proper scale parameter $\sigma^2$ facilitates the detection of various sized salient regions.

The qualitative comparison is shown in \figref{fig:visualcomp}.
The saliency maps of $\text{SDS}_\text{GM}$ are used to represent the proposed model.
As shown, the saliency maps of the recent models are more visually consistent with the ground-truth fixation density maps.
By contrast, the early works tend to highlight the local distinctive regions.
Moreover, among all the saliency models, although $\text{SDS}_\text{GM}$ is the most stable and accurate to predict the fixations in the examples,
it only computes the global distinctiveness of the simple feature: GM.
The portrait images in the last two columns of \figref{fig:visualcomp} are still complicated cases for $\text{SDS}_\text{GM}$.

{
\subsection{Comparison with Other Normalization Approaches}
\label{sec:norm}
From the perspective of normalization, the features in the proposed model is normalized by the patch-based global comparison of relative difference (i.e., the global structural dissimilarity),
whereas the features in Itti's model is normalized by the MMLM.
By computing the difference between the maximum and the mean of local maxima, MMLM can promote the feature maps with a small number of strong peaks.
However, in contrast to the proposed measure, MMLM cannot capture the difference of local structures to which the HVS is highly sensitive.

For comparison, according to Itti's model,
the proposed model based on \equref{equ:saliencyS} can be modified as
\begin{equation}
\label{equ:itti}
S_{map} = \mathcal{G}\left(\sum_{i=1}^3\mathcal{N}(\bm{f}_i)\right).
\end{equation}
In \equref{equ:itti}, compared with \equref{equ:saliencyS},
the patch-based global comparison of relative difference is replaced by the MMLM-based normalization (denoted by $\mathcal{N}$); the exponential operation is removed; the Gaussian blurring $\mathcal{G}$ with the standard deviation set as $5\%$ of the width of $S_{map}$ is conducted; and $\bm{f}_i$ is set as the LC feature.

In addition to MMLM, another two widely used normalization approaches, i.e., the 0-1 normalization and the graph-based approach \cite{Harel2006}, are adopted as alternatives of $\mathcal{N}$. The 0-1 normalization scales the feature maps to a fixed dynamic range $[0,1]$.  The graph-based normalization also computes the spatially weighted global difference, and is conducted on the value of features.

The models based on the three different normalization approaches are tested on the Toronto database.
The results shown in \tabref{tab:norm} support the claim that the proposed approach is better than other normalization approaches in the highlighting of the salient regions.


\begin{table}
\centering
\caption{Performance comparison of the model based on \equref{equ:itti} with different normalization approaches and the proposed $\text{SDS}_\text{LC}$.
}
\label{tab:norm}
\begin{tabular}{lcccc}
\toprule
$\mathcal{N}$& EMD & AUC & CC & NSS\\
\toprule
0-1  & 3.275 & 0.780 &0.439 &1.100\\
MMLM \cite{Itti1998}  & 2.274 & 0.781 &0.442 &1.109\\
graph \cite{Harel2006} &2.794 & 0.792 &0.519 &1.345\\
\midrule
$\text{SDS}_\text{LC}$  & 1.753 & 0.841 & 0.688 & 1.829\\
\bottomrule
\end{tabular}
\end{table}
}

\subsection{Complexity Analysis}
{
The proposed model involves two key stages, i.e., the structural feature extraction and the patch based global relative difference computation.
In the first stage, the bicubic interpolation of pixels based image resizing and the pixel-based RGB$\slash$YIQ should be conducted on an image first. Then, the convolution-based gradient computation or Gaussian filtering is needed to obtain the GM or LC features.
Note that the number of pixels in the resized image is $M\!w\!\times\!N\!w$ (denoted by $\!\Omega$).
Thus, the time complexities of image resizing, color space conversion, and convolution are all $O(\!\Omega\!)$.
In the second stage,
the salience of each image patch is computed using \equref{equ:saliencyS}.
Specifically, each patch is compared with other patches point-by-point using \equref{equ:newD} in three feature channels.
Thus, it requires $3\!\times\!\Omega$ times calculation of \equref{equ:newD}, $3\!\times\!\Omega\!-\!1$ times of additions, $\Omega$ times of multiplications, and one exponential operation to compute the salience of a patch.
Thus, when computing the salience of all the patches, the total time complexity of the second stage is $O(\Omega\times({M\!\times\!N}))$.
}

\begin{table}
\centering
\caption{Average Running Time over the Three Databases. All the competing models are implemented in Matlab, C, or Matlab + C.}
\label{tab:time}
\begin{tabular}{lcc}
\toprule
Models & Implementation & Running time (s)\\
\toprule
IT & M+C & 0.23\\
GBVS& M+C & 0.37\\
JUDD& M+C & 12.07 \\
LG & M+C & 0.59\\
CA & M & 32.70\\
SWD & M & 1.00\\
FES & M & 0.18\\
HFT & M & 0.15\\
SSD & M & 0.04\\
BMS & C & 0.05\\
LDS & M & 0.59\\
\midrule
$\text{SDS}_\text{LC}$ & M &0.57\\
$\text{SDS}_\text{GM}$ & M &0.57\\
\bottomrule
\end{tabular}
\end{table}

\tabref{tab:time} shows the running time comparison between the proposed models and the other 11 saliency models.
All the models were run on a computer with 4.00 GHz CPU and 16G RAM.
With Matlab implementation, the proposed $\text{SDS}_\text{LC}$ and $\text{SDS}_\text{GM}$ take approximately 0.5 s to generate a saliency map based on the average time cost on the three databases.
The running time of the proposed models was moderate in all the competing models.
The low running time is because all the operations except the feature extraction in the proposed framework are conducted in a point-wise manner.

\begin{figure}
    \centering
    \includegraphics[width=0.8\linewidth]{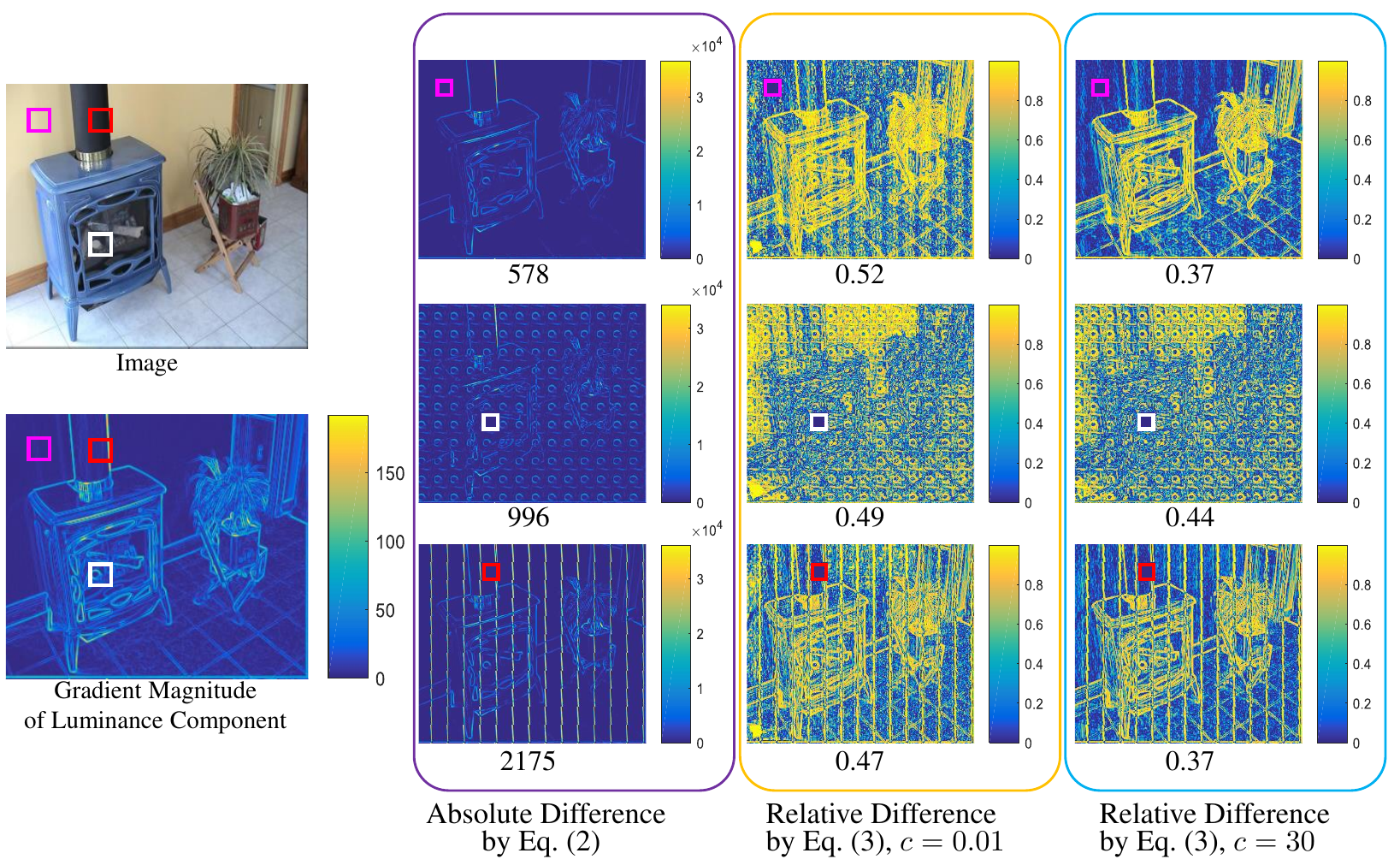}
    \caption{Illustration of the defects of the absolute difference. The first row is the example image and its GM map of luminance. Three patches are selected from the GM map, of which the purple patch is obtained from the insignificant background, the red patch possesses sharp and sparse boundaries, and the white patch is obtained from the inner region of the salient object. The selected patches are compared with the other patches point-by-point through different dissimilarity measures. The second--fourth rows show the difference maps and the corresponding average differences of different measures.
    }
    \label{fig:AbsoluteIsBad}
\end{figure}

{
\section{Analysis of the Proposed SDS Model}
\label{sec:analysis}
In this section, we seek to identify the reasons for the good performance of the proposed model in terms of three aspects: 1) the framework (i.e., spatially weighted dissimilarity); 2) the structural features; 3) the correlation-deduced relative difference.
Specifically, aspects 2) and 3) constitute the proposed structural dissimilarity measure.

\subsection{Framework}
The proposed model is based on a standard spatially weighted dissimilarity-based framework, and thus, it has the same advantage as other models based on this framework.
Specifically, the spatial weighting implicitly assigns larger weights to the center patches \cite{Harel2006}.
The implicit center bias is significant in addressing the photographer bias and the central fixation tendency of the observers \cite{tatler2007central}.
For the proposed model, if the structural dissimilarity is pruned (i.e., the dissimilarity $\bm{D}^x(p)$ is directly set as one in \equref{equ:saliencyS}), the remainder is the implicit center bias.
Based on the experiments, the corresponding NSS scores of the remainder model (1.274, 1.279, and 1.225 on the Toronto, MIT, and ImgSal databases, respectively) are good.
Therefore, the first reason for the good performance of the proposed model is that the adopted framework involves an implicit center bias.

\subsection{Structural Features}
The proposed dissimilarity is measured based on the structural features, which are extracted by the GM/LC in the YIQ color space. The structural features are crucial for the proposed model, because an intensity-color-opponent-based (ICO) color space  can highlight the salient regions with distinct colors. Then, the structure extractors GM or LC can make the salient regions stand out further.

To verify the effects of the structural features,
the YIQ color space was replaced by each of YCbCr and RGB, of which the YCbCr is also an ICO color space; the structure extractor was disabled
so that the dissimilarity could be directly measured based on the colors.
In the experiments, GM was used as the representative structure extractor.
The results obtained on the Toronto database are summarized in \tabref{tab:feat}.
When the GM is enabled, the proposed model shows similar performance on the YIQ and YCbCr color space. However, the performance on the RGB color space is obviously inferior. When the GM is disabled, the performance declines significantly for all the color spaces. Therefore, we can conclude that both the structure extractors and color space are important for the proposed model.

\begin{table}
\centering
\caption{Performance comparison of the proposed $\text{SDS}$ with features extracted in different color space and GM enabled$\slash$disabled ($+\slash-$). Specifically, $\text{SDS}_\text{YIQ+GM}$ is the original $\text{SDS}_\text{GM}$.}
\label{tab:feat}
\begin{tabular}{lcccc}
\toprule
& EMD & AUC & CC & NSS\\
\toprule
$\text{SDS}_\text{YCbCr+GM}$  & 1.758 & 0.843 &0.687 &1.826\\
$\text{SDS}_\text{YCbCr-GM}$  & 2.136 & 0.820 &0.579 &1.479\\
$\text{SDS}_\text{RGB+GM}$   &1.834 &0.837 &0.641 &1.623\\
$\text{SDS}_\text{RGB-GM}$   &3.267 &0.750 &0.403 &0.971\\
$\text{SDS}_\text{YIQ-GM}$   &2.013&0.826&0.608&1.558\\
\midrule
$\text{SDS}_\text{YIQ+GM}$  & 1.735&0.844&0.694&1.842\\
\bottomrule
\end{tabular}
\end{table}

\subsection{Correlation-deduced relative difference}
In contrast to the absolute difference, the relative difference can limit the value of the difference.
Therefore, the global difference (i.e., saliency as in \equref{equ:saliencyS}) will not be greatly affected by individual excessive differences.
The dissimilarity in the organization of the structural features is then highlighted.
\figref{fig:AbsoluteIsBad} illustrates the residue maps computed based on the absolute difference and relative difference for three example patches.
The mean difference is shown under the corresponding residue map.
}
The red-marked patch only possesses small area of boundaries.
However, the sharp boundaries still lead to a large average absolute difference.
By contrast, the difference is limited by the relative difference measure.
\figref{fig:AbsoluteIsBad} also shows the importance of the normalization parameter $c$.
Although the relative difference can limit the large absolute difference,
the difference between insignificant features would simultaneously be enhanced.
For example, when $c=0.01$, the background patch (marked by purple box) shows the largest mean relative difference.
However, when $c=30$, the differences between the insignificant features are obviously suppressed.
Specifically, an obvious change occurs in the background regions.

To further validate the advantage of the correlation deduced relative difference,
we replaced the relative dissimilarity measure (\equref{equ:newD}) by the absolute difference (\equref{equ:mse}) in the proposed model.

The corresponding model with tuned parameters is tested on the Toronto database.
The results evaluated on the Toronto database are 2.350, 0.818, 0.561, and 1.453 in terms of EMD, AUC, CC, and NSS, respectively. It can be seen that the scores are significantly reduced compared with the results of the original model as in \tabref{tab:allperformance}.
Therefore, we can conclude that the relative difference significantly improves the performance of the proposed model in saliency detection compared with the normally used absolute difference.

Based on the above analysis, all the three aspects attribute to the good performance of the proposed model.

\section{{Discussion and Conclusion}}
\label{sec:conclusion}
In this study, the image quality assessment (IQA) was introduced into saliency detection.
We proposed the use of the structural dissimilarity induced by the IQA models as the dissimilarity measure between image patches.
To highlight the effectiveness of the proposed structural dissimilarity,
we adopted a simple spatially weighted dissimilarity framework that excludes the multi-scale operation or post-processing.
Compared with the other 11 state-of-the-art saliency models, the proposed model presented highly competitive saliency detection performance tested on three saliency databases.
{Based on the results of the comprehensive experiments in Section~5, the structural features in the color space and the relative difference based measure (i.e., the proposed structural dissimilarity) attribute to the high performance of the proposed model. Based on the computation of the spatially weighted structural dissimilarity, the regions with the distinctive structures are highlighted.
According to the biological theories \cite{treisman1980feature,desimone1995neural}, image regions that stand out from their background are prioritized at almost all levels of the visual system and will attract human attention. Therefore, the highlighting of the structurally distinctive regions based on the proposed model is consistent to the allocation of human fixation.

Besides, the fixation map is generated by summing up the fixation points of a number of viewers during eye scanning on an image. In most works of saliency prediction including ours, the benchmark is the fixation map and not the eye scanpaths. In the works such as Ref.~\citenum{Itti1998}, the inhibition-of-return scheme is adopted as the post-processing on saliency map to predict the eye scanpaths on an image. Then, the eye scanpaths are predicted according to the descending salient values. However, besides the salient values, the eye scanpaths also depend on the context of the image content and the prior information \cite{Friston2012}. Therefore, an updating saliency map between saccades considering such a dependence of context and prior is expected to benefit the prediction of eye scanpaths. In future works, we will continue to investigate the prediction of eye scanpaths.
}


\bibliography{JEI-YangLi-arxiv}   
\bibliographystyle{spiejour}   

\listoffigures
\listoftables

\end{spacing}
\end{document}